\documentclass[
 reprint,
 superscriptaddress,
 amsmath,
 amssymb,
 pre
]{revtex4-1}

\usepackage{graphicx}% Include figure files
\usepackage{dcolumn}% Align table columns on decimal point
\usepackage{bm}% bold math

\usepackage{framed} % boxed paragaraph
\usepackage{placeins} % for \FloatBarrier
\usepackage{algpseudocode}
\usepackage{algorithm}
\usepackage{mathtools} % for \coloneq

\newcommand{\loss}{\mathcal{L}}

\begin{document}

%\preprint{APS/123-QED}

\title{Growing Neural Networks: Dynamic Evolution through Gradient Descent}

\author{Anil Radhakrishnan}
\affiliation{Nonlinear Artificial Intelligence Laboratory, Physics Department, North Carolina State University, Raleigh, NC 27607, USA }

\author{John F. Lindner*}
\affiliation{Nonlinear Artificial Intelligence Laboratory, Physics Department, North Carolina State University, Raleigh, NC 27607, USA }
\affiliation{Physics Department, The College of Wooster, Wooster, OH 44691, USA}

\author{Scott T. Miller}
\affiliation{Nonlinear Artificial Intelligence Laboratory, Physics Department, North Carolina State University, Raleigh, NC 27607, USA }

\author{Sudeshna Sinha}
\affiliation{Indian Institute of Science Education and Research Mohali, Knowledge City, SAS Nagar, Sector 81, Manauli PO 140 306, Punjab,
India}

\author{William L. Ditto}
\affiliation{Nonlinear Artificial Intelligence Laboratory, Physics Department, North Carolina State University, Raleigh, NC 27607, USA }

\date{\today}

\begin{abstract}
In contrast to conventional artificial neural networks, which are structurally static, we present two approaches for evolving small networks into larger ones during training. The first method employs an auxiliary weight that directly controls network size, while the second uses a controller-generated mask to modulate neuron participation. Both approaches optimize network size through the same gradient-descent algorithm that updates the network's weights and biases. We evaluate these growing networks on nonlinear regression and classification tasks, where they consistently outperform static networks of equivalent final size. We then explore the hyperparameter space of these networks to find associated scaling relations relative to their static counterparts. Our results suggest that starting small and growing naturally may be preferable to simply starting large, particularly as neural networks continue to grow in size and energy consumption.
\end{abstract}

%\keywords{Suggested keywords}%Use showkeys class option if keyword
                              %display desired
\maketitle

%\tableofcontents

\section{Introduction}
\label{sec:intro}
The exponential growth in artificial neural network size has become a critical concern in machine learning. Modern language models have grown from millions of parameters to hundreds of billions in just a few years,
with GPT-3 containing 175 billion parameters and more recent models exceeding a trillion parameters~\cite{brown2020language}. This growth comes with substantial computational and environmental costs --- training a single large language model can emit as much carbon as five cars over their entire lifetimes~\cite{strubell2019energy}, and the energy consumption of neural-network training runs has been doubling every 3.4 months~\cite{openai2018ai}.

Despite this trend toward larger models, evidence suggests that many neural networks are significantly overparameterized for their tasks~\cite{frankle2018lottery}. Studies have shown that up to 90\% of parameters in some networks can be pruned after training with minimal impact on performance~\cite{see2016compression}. This raises a fundamental question: rather than starting with large networks and pruning them down, could we start small and grow networks to their optimal size during training?

Natural systems provide inspiration for this approach. Biological neural networks, from C. elegans to human brains, grow and prune connections throughout development~\cite{zhao2024integrative}. This dynamic architecture allows them to achieve remarkable computational efficiency; the human brain performs complex cognitive tasks while consuming only about 20 watts of power~\cite{gebickehaerter2023computational}.

Beyond neural systems, dynamic network growth is ubiquitous in nature and technology. Social networks evolve through the addition of new connections~\cite{fan2004evolving}, transportation networks expand to meet changing demands~\cite{aggarwal2014evolutionary}, and disease transmission networks adapt during epidemics~\cite{tunc2014effects}. These networks share a common feature: they grow and adapt in response to functional demands rather than starting at maximum size.

Current approaches to neural network architecture optimization largely fall into three categories: (1) static networks with hand-designed architectures, (2) neural architecture search methods that explore fixed architectures~\cite{elsken2018neural}, and (3) post-training pruning techniques that remove unnecessary connections~\cite{blalock2020what}. While these methods have shown success, they either require significant computational resources for architecture search or maintain unnecessarily large networks during training.

Several attempts have been made to develop growing neural networks. The Neuro-Evolution of Augmenting Topologies~(NEAT) algorithm uses genetic algorithms to evolve network structures~\cite{stanley2002evolving}, while cascade correlation networks add neurons incrementally during training~\cite{fahlman1989cascade}. However, these methods either do not leverage powerful gradient-based optimization or require complex evolutionary algorithms.

In this paper, we present a novel approach to neural network growth that integrates seamlessly with gradient-based optimization. Our key innovation is to make network size itself a differentiable parameter that can be optimized alongside weights and biases. We present two complementary implementations: an auxiliary-weight algorithm that explicitly controls network size through a single learnable parameter, and a controller-mask approach that provides more scalable and efficient control over network growth.

We demonstrate that our growing networks can consistently outperform static networks of equivalent final size on both regression and classification tasks. This performance advantage appears to stem from reduced susceptibility to local minima during training, as smaller networks naturally have simpler loss landscapes. Furthermore, our approach theoretically requires significantly less computation during early training stages when networks are small.

The remainder of this paper is organized as follows: Section~\ref{sec:theory} presents the theoretical framework for gradient-based network growth, Section~\ref{sec:impl} details our two implementation approaches, Section~\ref{sec:res} presents experimental results, and Section~\ref{sec:dis} discusses implications and future directions.

\section{Theoretical Framework}
\label{sec:theory}

Feed-forward neural networks are composed of layers of interconnected nodes, transforming input data $x$ through a series of nonlinear operations. For a network with $L$ layers, the output $\hat{y}$ can be expressed as
\begin{equation}
    \hat{y}(x) = f_L(f_{L-1}(\cdots f_1(x))),
\end{equation}
where each layer's transformation is
\begin{equation}
    f_l(x) = \sigma(W_lx + b_l),
\end{equation}
where $W_l$ represents the weight matrix, $b_l$ the bias vector, and $\sigma$ the activation function for layer $l$. Traditional training minimizes a loss function $\loss$ through gradient descent,
\begin{equation}
    \theta_{t+1} = \theta_t - \eta \nabla_\theta \loss(\theta_t),
\end{equation}
where $\theta$ represents all trainable parameters and $\eta$ is the learning rate.

Our key innovation is to make network size itself a differentiable parameter that can be optimized during training. We augment the standard loss function with a size-dependent term
\begin{equation}
    \loss = \loss_{\text{base}} + \lambda \loss_{\text{size}},
\end{equation}
where $\loss_{\text{base}}$ measures performance on the target task (e.g., mean squared error for regression), $\loss_{\text{size}}$ tunes network complexity, and size-influence parameter $\lambda$ balances these objectives. The size-dependent loss can take various forms, such as
\begin{equation}
   \loss_{\text{size}} = (N - N_{\text{target}})^2,
   \label{eq:size_loss}
\end{equation}
where $N$ is the current network size and $N_{\text{target}}$ is the desired final size. This quadratic penalty creates a smooth optimization landscape for network growth.

Network growth inherently involves discrete architectural changes. For a network with $N$ potential neurons, each neuron's state can be viewed as a binary decision: active $(1)$ or inactive $(0)$. This creates a discrete search space of size $2^N$, which is not directly amenable to gradient-based optimization. The key challenge is to transform these discrete architectural decisions into a continuous, differentiable form that can be optimized using gradient descent while maintaining the network's functionality during training.

Two general strategies exist for creating a differentiable approximation to discrete network growth:
\begin{enumerate}
    \item \textbf{Structural modification:} One can directly modify the network's structure through continuous parameters that control the effective presence or absence of neurons. This requires careful design of differentiable transitions between network sizes.
    \item \textbf{Activity modulation:} Alternatively, one can maintain a fixed maximum-size network but continuously control the contribution of each neuron through differentiable scaling factors.
\end{enumerate}
Both approaches require a smooth transition function $\psi(x)$ that approximates a step function while maintaining differentiability. We choose
\begin{equation}
    \psi(x) = \begin{cases} 
    1, & x < -1, \\
    \sin^2(\frac{\pi}{2}x), & -1 \leq x \leq 0, \\
    0, & x > 0.
    \end{cases}
    \label{eq:trans}
\end{equation}
This allows delicate control over the size and for gradients to flow through what would otherwise be discrete architectural decisions. In either approach, the network's output can be generally expressed as,
\begin{equation}
    \hat{y}(x) = f(x; \theta, \alpha),
\end{equation}
where $\theta$ represents the standard network parameters (weights and biases), and $\alpha$ represents the continuous parameters controlling network size or neuron participation. The total loss function then becomes
\begin{equation}
    \loss = \loss_{\text{base}}(f(x; \theta, \alpha), y) + \lambda \loss_{\text{size}}(\alpha).
\end{equation}
This formulation allows joint optimization of network parameters and size through the gradient descent
\begin{align}
    \frac{\partial \loss}{\partial \theta} &= \frac{\loss_{\text{base}}}{\partial \theta},\\
    \frac{\partial \loss}{\partial \alpha} &= \frac{\loss_{\text{base}}}{\partial \alpha} + \lambda \frac{\partial  \loss_{\text{size}}}{\partial \alpha}.
\end{align}
While these gradient relationships hold for standard optimizers; for adaptive optimizers like Adam, the size influence parameter $\lambda$ becomes redundant due to automatic gradient scaling (see Appendix~\ref{app:adaptive}).

The transition function $\psi(x)$ plays a crucial role in the gradient flow through $\partial f/\partial \alpha$. For the piecewise continuous form in equation~\ref{eq:trans}, the derivative in the transition region is
\begin{equation}
    \frac{\partial \psi}{\partial x} = 
    \begin{cases}
        0, & x < -1, \\
        \pi \sin(\frac{\pi}{2}x)\cos(\frac{\pi}{2}x), & -1 \leq x \leq 0, \\
        0, & 0 < x.
    \end{cases}
\label{eq:trans_der}
\end{equation}
This smooth derivative ensures stable gradient flow during size transitions, while the zero derivatives in the saturated regions help maintain stability when neurons are fully active or inactive.

\begin{figure*}[htb]
    \centering
    \includegraphics[width=0.9\linewidth]{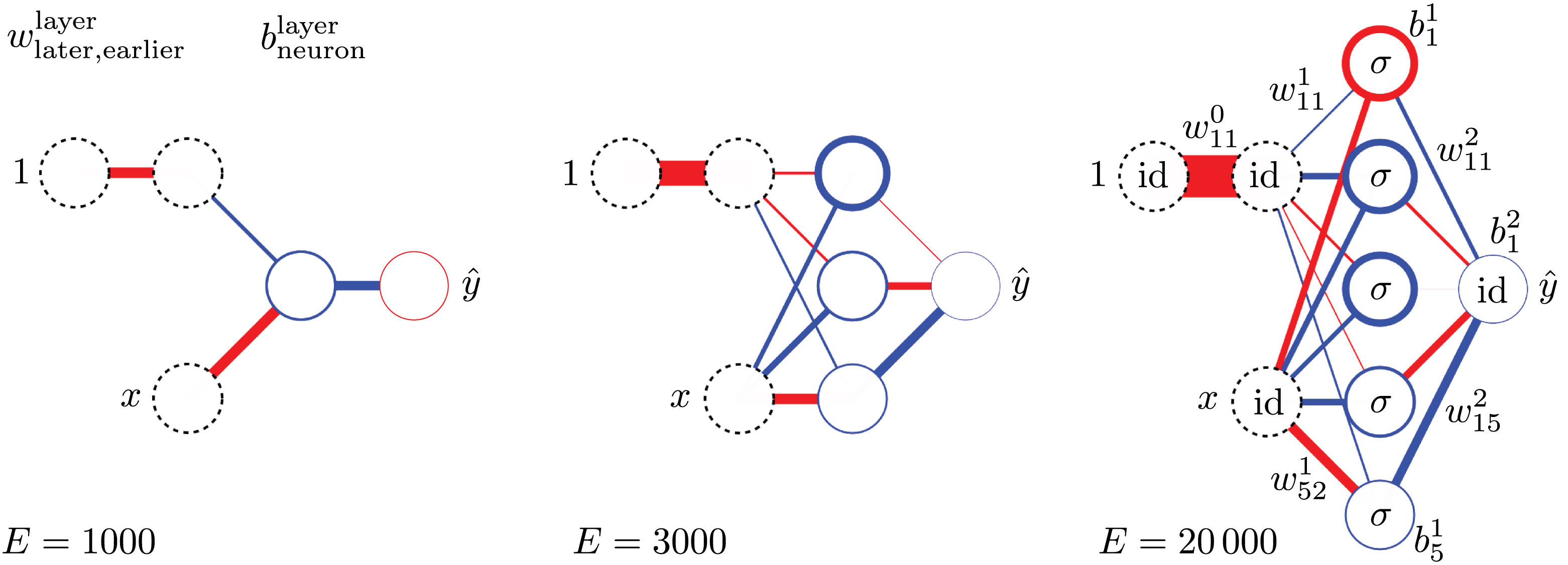}
    \caption{Loss function drives network size from 0 to 5 hidden neurons via gradient descent. At each training round or epoch $E$, lines represent weights and circles biases, partially labeled at bottom, thicknesses are proportional to magnitudes, red is positive and blue is negative. The prepended $1$ is converted to the weight $w^0_{11} = a^0_1$ by an identity activation with a zero bias, and $\smash{N = w^0_{11}}$ is the network size, which gradient descent naturally adjusts along with the other weights and biases.}
    \label{fig:aux_wt_scheme}
\end{figure*}

\begin{figure*}[htb]
    \centering
    \includegraphics[width=0.85\linewidth]{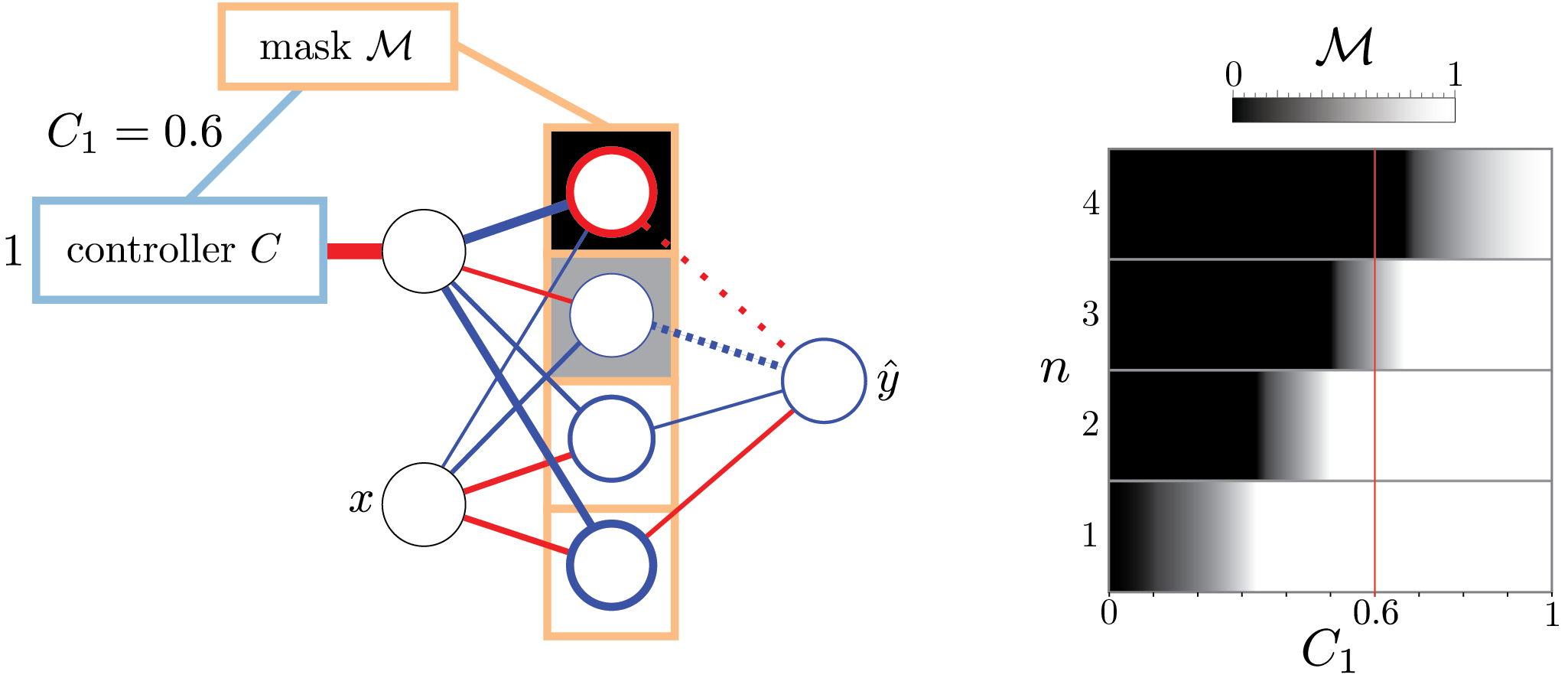}
    \caption{Controller-mask paradigm schematic for up to $N=4$ hidden neurons. Mask is mostly open with $C_1 = 0.6$, two hidden neurons ``on'' (white squares), one partially ``on'' (grey square), and one ``off'' (black square). Lines represent weights and circles biases, thicknesses are proportional to magnitudes, red is positive and blue is negative, dashed lines suggest the effects of multiplicatively masking neuron outputs.}
    \label{fig:controller_mask_scheme}
\end{figure*}

The convergence properties of our growing network approach can be analyzed by considering both $\loss_{\text{base}}$ and $\loss_{\text{size}}$ components. Building on classical neural network optimization theory~\cite{bottou2016optimization}, the loss function $\loss$ must satisfy:
\begin{enumerate}
    \item \textbf{$\beta$-smoothness}: For all $\theta_1, \theta_2, \alpha_1, \alpha_2$,
    \begin{equation}
        \|\nabla \loss(\theta_1, \alpha_1) - \nabla \loss(\theta_2, \alpha_2)\| \leq \beta\|(\theta_1, \alpha_1) - (\theta_2, \alpha_2)\|.
    \end{equation}

    \item \textbf{Lower boundedness}: For all $\theta,\alpha$,
    \begin{equation}
        \loss(\theta, \alpha) \geq \loss_{\text{min}} > -\infty.
    \end{equation}

    \item \textbf{Size-loss Convexity}: The size loss $\loss_{\text{size}}$ is convex in $\alpha$ due to its quadratic form
    \begin{equation}
        \loss_{\text{size}}(\alpha) = (N(\alpha) - N_{\text{target}})^2.
    \end{equation}
\end{enumerate}

The $\beta$-smoothness property is inherited from the neural network architecture with bounded activation functions~\cite{allen2019convergence}, while the transition function $\psi(x)$ ensures smoothness during size transitions. For learning rate $\eta< 2/\beta$, we can prove the descent lemma to find that the coefficient of $\|\nabla\loss\|^2$ is negative, giving
\begin{equation}
        \loss(\theta_{t+1}, \alpha_{t+1}) \leq \loss(\theta_t, \alpha_t) - \frac{\eta}{2}\|\nabla \loss\|^2.
\end{equation}
This can be naturally extended to show convergence in stochastic settings with mini-batch gradients~\cite{ghadimi2013stochastic}.

The transition function $\psi(x)$ also plays a crucial role in stability. We can show the network size changes are bounded by
\begin{equation}
    |N_{t+1} - N_t| \leq \eta\|\nabla_\alpha \loss\| \leq \eta M,
\end{equation}
where $M$ is bounded by the Lipschitz constant of $\psi$, and transition function changes are bounded by
\begin{equation}
    |\psi(x_{t+1}) - \psi(x_t)| \leq K\|x_{t+1} - x_t\|,
\end{equation}
 with Lipschitz constant $K$. These imply bounds on the learning rate
\begin{equation}
    \eta \leq \min\left\{\frac{2}{\beta}, \frac{\Delta N_{\text{max}}}{M}\right\},
\end{equation}
where $\Delta N_{\text{max}}$ is the maximum allowed size change per iteration~\cite{e2019comparative}.

\section{Implementation Approaches}
\label{sec:impl}

We present two distinct approaches to implementing dynamic network growth:
\begin{enumerate}
    \item The \textbf{auxiliary-weight} algorithm, which directly modifies network structure.
    \item The \textbf{controller-mask} algorithm, which modulates neuron participation.
\end{enumerate}

\subsection{Auxiliary-Weight Algorithm}

The auxiliary-weight algorithm achieves network growth by identifying network size with a learnable parameter in the network itself.

Consider a feed-forward neural network with one hidden layer. We prepend a layer $0$ that consists of a single input fixed at $1$ connected to the first layer through a weight $w_{11}^{0}$. This weight serves as our size parameter $N = w_{11}^{0}$, controlling the effective number of active neurons in the hidden layer. This technique of using an auxiliary weight to set neural network constraints was first used by Jin et al. in the context of physics-informed neural networks~\cite{jin2022physicsinformed}.

The network computation proceeds as:
\begin{enumerate}
\item Input Layer (Layer 0):
\begin{equation}
    N = w_{11}^0 = a_1^0 = \text{id}(w_{11}^0 \cdot 1 + 0),
\end{equation}
where the prepended 1 is converted to $N$ through an identity activation with zero bias.

\item Hidden Layer (Layer 1):
\begin{equation}
    a_i^1 = \psi_{i-N}\sigma(w_{i1}^1 x + b_i^1),
\end{equation}
where $\psi$ modulates activation based on relative position to $N$.
   
\item Output Layer (Layer 2):
\begin{equation}
    \hat{y} = \sum_{i=1}^{N_{\text{max}}} w_{1i}^2 a_i^1 + b_1^2.
\end{equation}
\end{enumerate}
During training, the network size $N$ evolves alongside other parameters through gradient descent
\begin{equation}
    N_{t+1} = N_t - \eta \frac{\partial \loss}{\partial N}.
\end{equation}
As $N$ increases, new neurons smoothly activate and join the computation, as shown in figure~\ref{fig:aux_wt_scheme}.

\subsection{Controller-Mask Algorithm}

The controller-mask algorithm maintains a fixed-size network but controls neuron participation through a learnable mask filter.

The architecture consists of two main components:
\begin{enumerate}
    \item A standard \textbf{multilayer perceptron}~(MLP) with fixed maximum size.
    \item A \textbf{controller} that generates masking values.
\end{enumerate}
The controller outputs a value  $C_1 \coloneq  C(1)$ that determines the effective network size,
\begin{equation}
    \tilde N = N\sin^2\left(\frac{\pi}{2}C_1\right).
\end{equation}
%
%For each neuron $n$, this generates the mask
The output of neuron $n$ is masked by
\begin{equation}
    \mathcal{M}(C_1,n) = \begin{cases}
1, & n < \lfloor \tilde N \rfloor, \\
\tilde N - \lfloor \tilde N \rfloor, & n = \lfloor \tilde N \rfloor, \\
0, & n > \lfloor \tilde N \rfloor,
\end{cases} 
\end{equation}
which is plotted in figure~\ref{fig:controller_mask_scheme} and is equivalent to the  equation~\ref{eq:trans} transition function $\psi$. While the controller can be arbitrarily complex, we can recover the simple single parameter control of the auxiliary-weight algorithm by defining
\begin{equation}
    C(x)= w\cdot x,
\end{equation}
where $C$ is the size controller operator, and $w$ is a tunable parameter. More generally, since $C_1$ is a normalized output, the size loss
\begin{equation}
    \loss_{\text{size}} = (C_1 - 1)^2
\end{equation}
is equivalent to equation~\ref{eq:size_loss}. The complete algorithm using this scheme is outlined in algorithm~\ref{Alg:controller_mask} and illustrated in figure~\ref{fig:controller_mask_scheme}.

\begin{algorithm}[H]
    \caption{Controller-mask grows an MLP while solving a regression problem.}
    \begin{algorithmic}[1]
        \Require Training data $(X_{\text{train}}, Y_{\text{train}})$, Number of epochs $E$, Learning rate $\eta$, Maximum neurons per hidden layer $N$, Size-loss coupling $\lambda$
        \Ensure Trained MLP model with dynamic neuron adjustment
        
        \State Initialize MLP model $M$ with input size $d_{\text{in}}$, output size $d_{\text{out}}$, and hidden layers $[h_1, h_2, \ldots, h_L]$
        \State Initialize Controller $C$
        \State Initialize Optimizer $\mathbb{O}(C, M)$
        % C_new = O(C)
        \For{epoch $= 1$ to $E$}
            \State $C_1 \gets C(\mathbf{1})$ \Comment{Compute control value}
            \State $X_{\text{new}} \gets \text{concatenate}(X_{\text{train}}, C_1)$ \Comment{Augment input with control value}
            
            \For{each layer $l$ in $M$}
                \State $\mathcal{M} \gets \text{control\_to\_mask}(C_1, N)$ \Comment{Compute neuron mask}
                \State $X_{\text{new}} \gets \text{apply\_mask}(X_{\text{new}}, \mathcal{M})$ \Comment{Apply mask to layer output}
                \State $X_{\text{new}} \gets \sigma(l(X_{\text{new}}))$ \Comment{Pass through layer with activation}
            \EndFor
            
            \State $Y_{\text{pred}} \gets M(X_{\text{new}})$ \Comment{Compute model prediction}
            \State $\loss_{\text{base}} \gets \text{mean}((Y_{\text{pred}} - Y_{\text{train}})^2)$ \Comment{Compute base loss}
            \State $\loss_{\text{size}} \gets \text{mean}((C_1 - 1)^2)$ \Comment{Compute size loss}
            \State $\loss \gets \loss_{\text{base}} + \lambda\loss_{\text{size}}$ \Comment{Total loss}
            
            \State $\mathbb{O} \gets \text{update\_optimizer}(\mathbb{O}, \loss)$
            \State $C \gets \text{update\_controller}(C,\mathbb{O}, \loss)$ \Comment{Update controller parameters}
            \State $M \gets \text{update\_model}(M,\mathbb{O}, \loss)$ \Comment{Update model parameters}
        \EndFor
        
        \State \Return Trained MLP model $M$ and Controller $C$
    \end{algorithmic} \label{Alg:controller_mask}
\end{algorithm}

\begin{figure*}[!htb]
    \centering
    \includegraphics[width=0.95\linewidth]{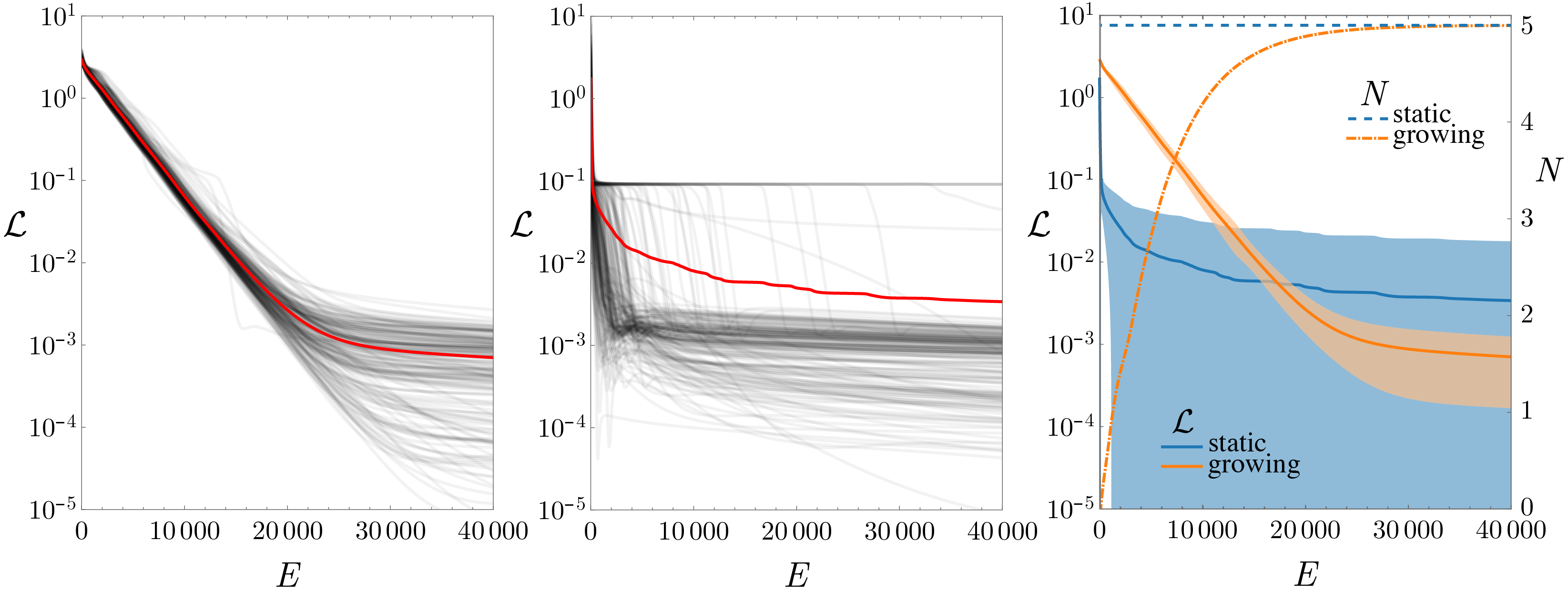}
    \caption{Auxiliary-weight algorithm nonlinear regression example. Training a growing network via a loss function (left) versus a known network (center) and combined with mean plus or minus standard deviation (right). Growing network outperforms static network, with the mean final static loss about 5 times the mean final growing loss. Trained using batch gradient descent with learning rate $\eta = 0.001$, and size-loss coupling $\lambda = 0.1$.}
    \label{fig:AuxNeuronRegression}
\end{figure*}

\subsection{Comparison}

The computational and memory requirements of the two approaches differ significantly in both their scaling behavior and practical implementation constraints. For the auxiliary-weight algorithm, the computational complexity of the forward pass scales with $\mathcal{O}(N_{\text{max}}  d)$ where $d$ is the input dimension, but the actual computation performed typically involves only $N_{\text{current}}$ active neurons, where $N_{\text{current}}\leq N_{\text{max}}$. This leads to efficient computation when the network is small, with computational costs growing as neurons activate. The memory requirements follow a similar pattern, scaling with $\mathcal{O}(N_{\text{current}}  d)$, making this approach particularly memory-efficient during early training phases.

In contrast, the controller-mask algorithm maintains a fixed computational graph of size $\mathcal{O}(N_{\text{max}}  d)$ throughout training. While this might seem less efficient, it enables better hardware utilization through parallel computation and can be more efficient on modern GPU architectures that prefer regular computation patterns. The memory footprint remains constant at $\mathcal{O}(N_{\text{max}}  d)$ plus a small overhead for the controller, making memory requirements more predictable but potentially larger than the auxiliary-weight approach for small networks.

The gradient computation complexity reflects these differences. The auxiliary-weight algorithm computes gradients only for active neurons and their connections, while the controller-mask algorithm computes gradients for all neurons but scales them by the mask values. This leads to an interesting trade-off: while the auxiliary-weight algorithm performs fewer computations overall, the controller-mask algorithm may execute faster on parallel hardware despite performing more total operations.

The training dynamics of growing networks present unique challenges that require careful consideration. In the auxiliary-weight approach, we initialize the size parameter $w_{11}^{0}$ near zero to begin with a minimal network. This initialization, combined with standard weight initialization for potentially active neurons, allows the network to grow naturally in response to task demands. The learning process typically exhibits a characteristic pattern: rapid initial growth as the network establishes basic task competency, followed by more gradual size adjustments as it refines its performance.

The controller-mask algorithm, while conceptually similar, approaches initialization differently. All network weights undergo standard initialization, but the controller is initialized to produce a mask that initially suppresses most neurons. This creates an effective small network within the larger architecture. The training process then modulates neuron participation through smooth adjustments to the mask values rather than structural changes.

We implemented the auxiliary-weight algorithm in Mathematica~\cite{Mathematica} and the controller-mask algorithm in Python using the JAX autodifferentiation framework~\cite{jax2018github} and Equinox neural network library~\cite{kidger2021equinox} with optimizers from Optax~\cite{deepmind2020jax}. Our code is available in our GitHub repository \footnote{Our code is available at our GitHub repository https://github.com/NonlinearArtificialIntelligenceLab/N3}.

\section{Experimental results}
\label{sec:res}
We test both the auxiliary-weight and the controller-mask algorithm on nonlinear regression and classification tasks finding regimes where the techniques show superiority over their conventional static counterparts.

\subsection{Nonlinear Regression}
For out regression task we choose to have the neural networks learn Bessel functions or their composition. 

The auxiliary-weight algorithm is tested on learning a simple Bessel function,
\begin{equation} \label{eq:bessel_simple}
    y(x) = a+b\,J_0(x),
\end{equation}
with $a$ and $b$ such that $y\in [-1,1]$ for $x\in [-1,1]$. This task serves as a proof of concept for learning with a growing network and using a batch gradient descent with  learning rate $\eta = 0.001$, and size-loss coupling $\lambda = 0.1$. The target network size is $n_\infty=5$ and the network is trained for $n_E = 4\times 10^4$ epochs with $n_t=40$ random data pairs where $80\%$ is used for training and $20\%$ for testing. 
The weights and biases are initialized from a uniform distribution with $w_{nm}^l, b_n^l \in [-1,1]$.

Figure~\ref{fig:AuxNeuronRegression} summarizes an auxiliary-weight algorithm nonlinear regression example. Training a growing network via a loss function (left) versus a static network (center) and combined with mean plus or minus one standard deviation (right). Growing networks outperform static networks averaged over $200$ trials. 

For the controller-mask algorithm we raise the complexity of the nonlinear regression problem by using a composition of Bessel functions,
\begin{equation} \label{eq:bessel_composite}
    y(x) = a + b(J_0(x)+J_1(x)+J_2(x)),
\end{equation}
with $a$ and $b$ such that $y\in [-1,1]$ for $x\in [-1,1]$. This task serves to strengthen our result for auxiliary-weight algorithm by by using a harder problem with more modern Adam adaptive optimizer with a learning rate $\eta = 0.001$.  The target network size is $n_\infty=10$ and the network is trained for $n_E = 5\times 10^3$ epochs with $n_t=2^{15}$ random data pairs where $80\%$ is used for training and $20\%$ for testing. 
The weights and biases are initialized from the normal distribution.

% \vspace{-0.5cm}
\begin{figure*}[tbh]
    \centering
    \includegraphics[width=0.9\linewidth]{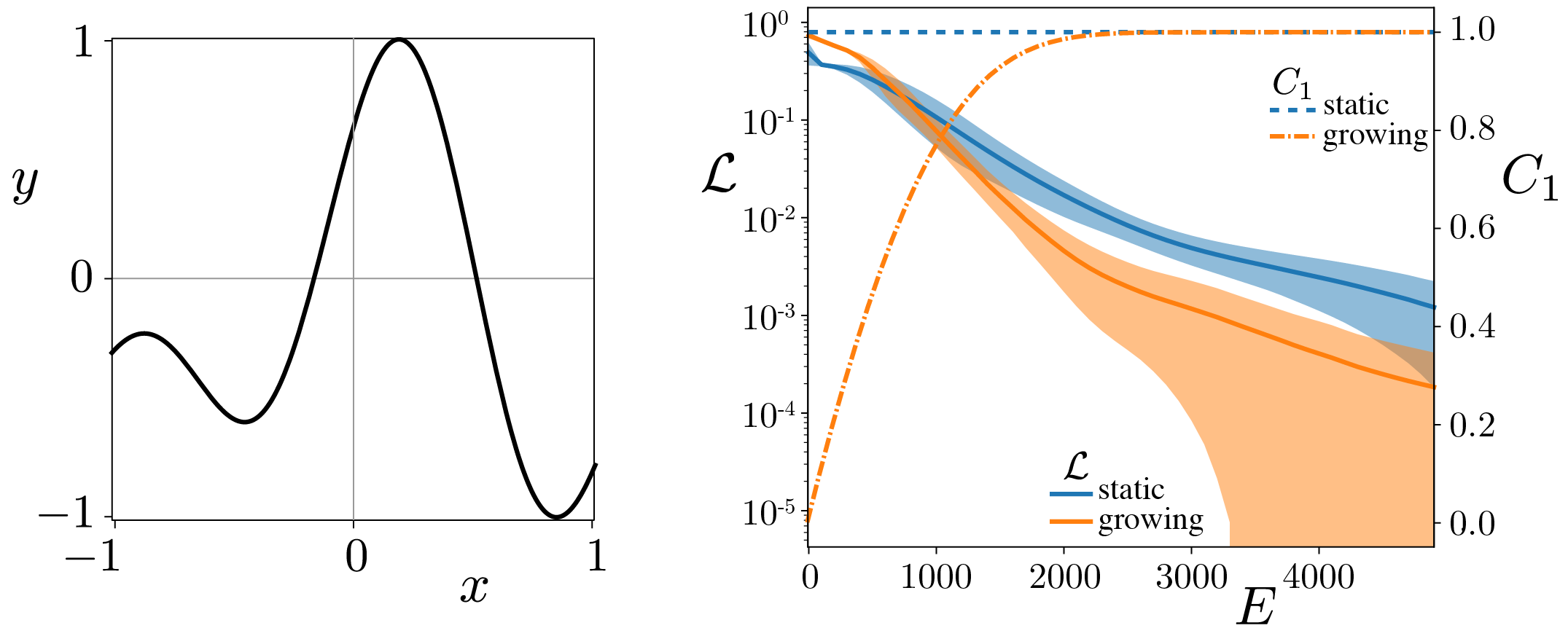}
%    \vspace{-0.5cm}
    \caption{Control-mask algorithm nonlinear regression example, target equation~\ref{eq:bessel_composite} composite Bessel function (left) and loss versus training epoch (right) for $2^{15}$ training pairs. Dark lines are mean test losses averaged over 100 trials and enclosing areas are plus or minus one standard deviation. In both cases, the growing network outperforms the static network. Trained using Adam optimizer with initial learning rate $\eta = 0.001$.}%, and size-loss coupling $\lambda = 0.32$.}
    \label{fig:ControlMaskRegression}
\end{figure*}

% \vspace{-1cm}

Figure~\ref{fig:ControlMaskRegression} summarizes a control-mask algorithm nonlinear regression example. Dark lines are mean test losses averaged over $100$ trials and enclosing areas are plus or minus one standard deviation. Growing networks outperform static networks just as with the auxiliary-weight algorithm.

\subsection{Classification}

For classification we chose the spiral dataset for its scalable complexity and broad popularity~\cite{chalup2007variations}.

\begin{figure*}[tbh]
    \centering
    \includegraphics[width=1\linewidth]{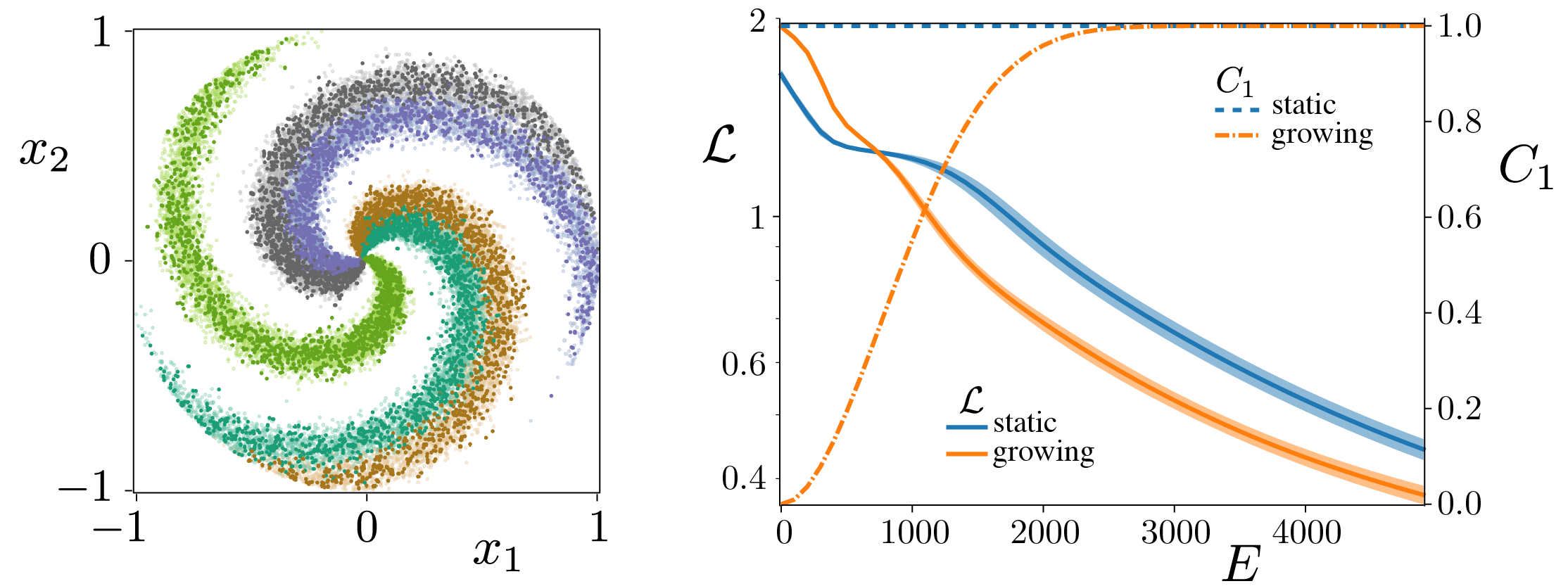}
    \caption{Control-mask algorithm classification example, target spirals (left) and loss versus training epoch (right) for $2^{15}$ training pairs. Dark lines are mean test losses averaged over 100 trials and enclosing areas are plus or minus one standard deviation. In both cases, the growing network outperforms the static network. Trained using Adam optimizer with initial learning rate $\eta = 0.001$.} %, and size-loss coupling $\lambda = 0.32$.}
    \label{fig:ControlMaskClassification}
\end{figure*}

Figure~\ref{fig:ControlMaskClassification} summarizes a control-mask algorithm classification example, with target spirals (left) and loss versus training epoch (right) for $2^{15}$ training pairs. Dark lines are mean test losses averaged over 100 trials and enclosing areas are plus or minus one standard deviation. In both cases, the growing network outperforms the static  network. Both networks were trained using the Adam optimizer with initial learning rate $\eta = 0.001$.%size-loss coupling $\lambda = 0.32$.

\subsection{Growth-Training Dependence}

\begin{figure*}[!htb]
    \centering
    \includegraphics[width=0.95\linewidth]{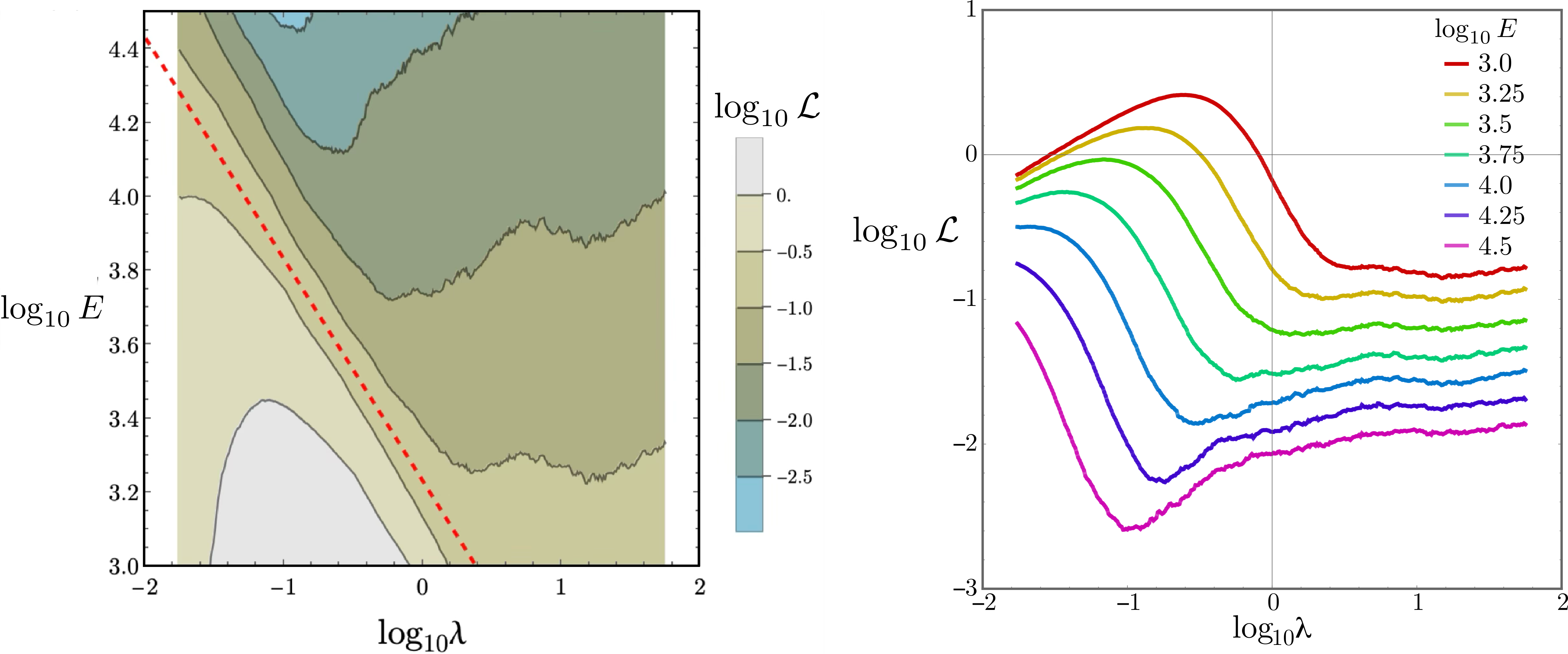}
    \caption{Performance dependence of auxiliary-weight algorithm for the equation~\ref{eq:bessel_simple} simple Bessel function nonlinear regression. Final Loss $\loss$ for growing networks across training duration $E$ and size-loss coupling $\lambda$. Smaller losses $\loss$ are better. Dashed line in density plot~(left) is approximately $E \approx 2000 \lambda^{-0.8}$}
    \label{fig:size_influence_loss}
\end{figure*}

\begin{figure*}[!htb]
    \centering
    \includegraphics[width=0.95\linewidth]{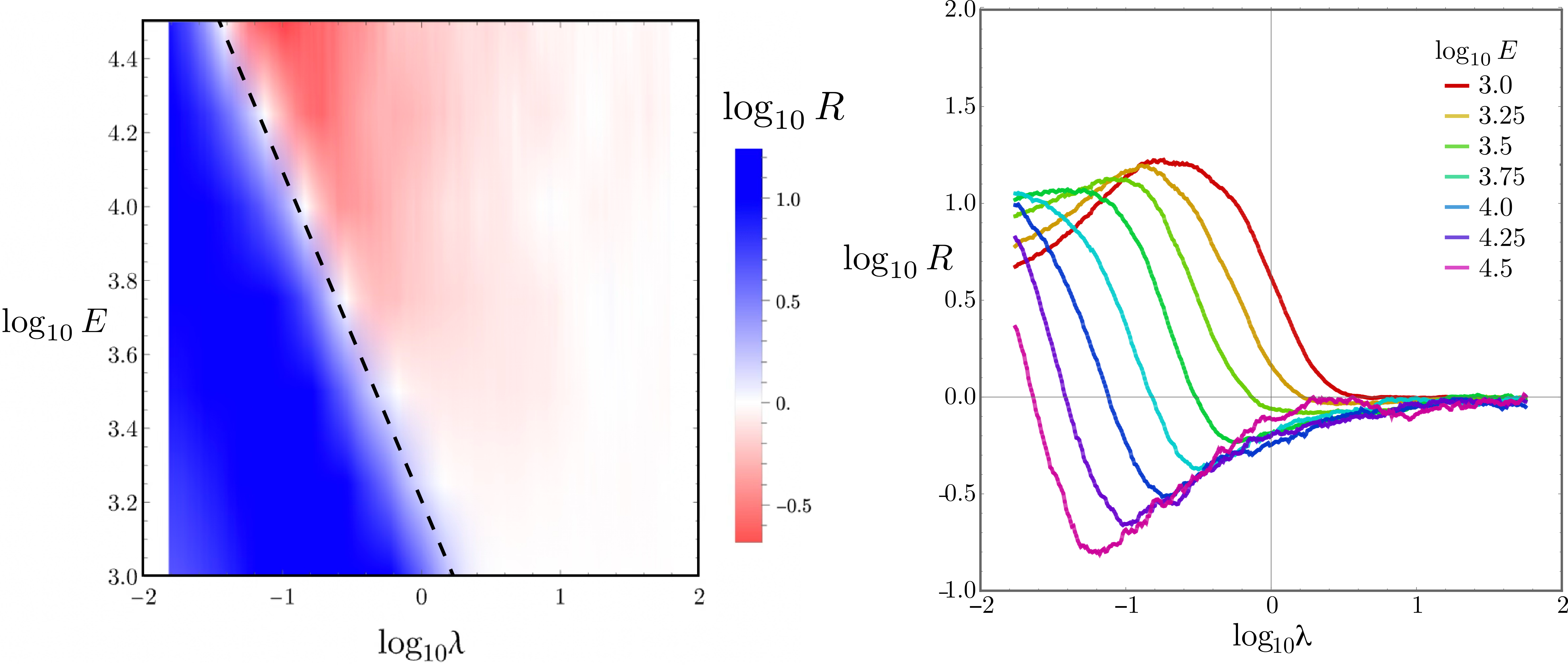}
    \caption{Relative Performance dependence of auxiliary-weight algorithm for the equation~\ref{eq:bessel_simple} simple Bessel function nonlinear regression against static networks trained for same duration $E$. Ratio $R=\loss^g_f/\loss^s_f$ of the final losses of growing networks over that of static networks across training duration $E$ and size-loss coupling $\lambda$. Ratios $R<1$ demonstrate growing network superiority. Dashed line in density plot~(left) is approximately $E \approx 2000 \lambda^{-0.8}$.}
    \label{fig:size_influence_ratio}
\end{figure*}

Figures~\ref{fig:size_influence_loss},~\ref{fig:size_influence_ratio} shows how the auxiliary-weight algorithm's performance depends on the training duration $E$ and size-loss coupling $\lambda$. We track the final loss $\loss$ for the growing networks and also the ratio $R=\loss^g_f/\loss^s_f$ of the final losses of growing networks over that of static networks.
Lower values of $\loss$ indicate better performance, and ratios $R<1$ demonstrate growing network superiority. We get similar results for both batch gradient descent (where weights and biases are updated after each epoch) and stochastic gradient descent (where weights and biases are updated after each training pair).

For large size-loss coupling $\lambda$, the growing network quickly grows to the same size as the static network, their final losses are similar, and the ratio $R\approx 1$. For small size-loss coupling $\lambda$, the growing network grows slowly and is still small at training's end, so its final losses are larger, and $R > 1$. Growing networks strategy is best for slow growth during long training. Linear features in density plots exhibit power law scaling, with dashed lines approximately $E \approx 2000 \lambda^{-0.8}.$

 \begin{figure*}[!htb]
    \centering
    \includegraphics[width=1.0\linewidth]{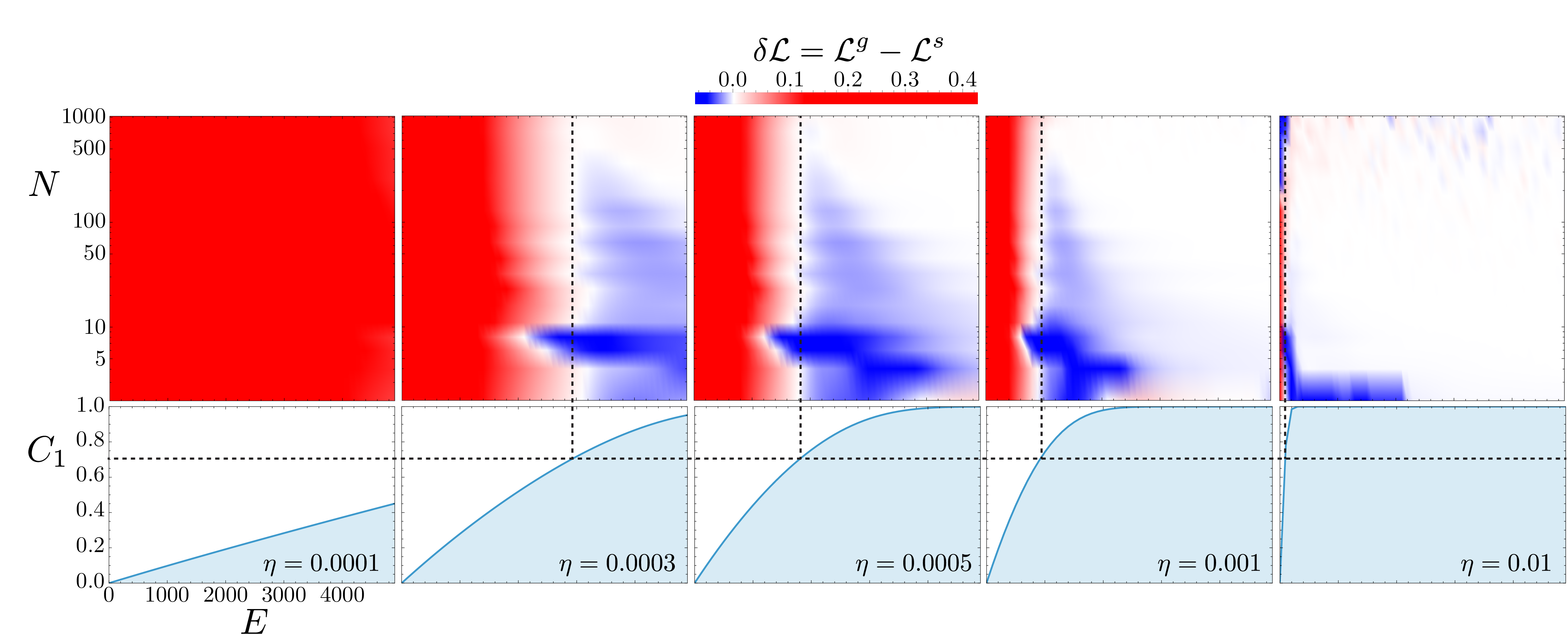}
    \caption{Scaling of controller-mask equation~\ref{eq:bessel_composite} composite Bessel function regression versus number of training epochs $E$ and hidden neurons $N$, for different learning rates $\eta$. Top row plots the difference $\delta \mathcal{L}$ between the mean final growing and static losses, where growing overperforms static in the blue regions. Bottom row plots the control value $C_1$ versus epoch. In between, when the growing network matures just at training's end, growing overperforms static. Dashed lines highlight transition at $C_1 \sim 0.7$.}
    \label{fig:controllerAlgBesselScaling}
\end{figure*}

Figure~\ref{fig:controllerAlgBesselScaling} reveals how the controller-mask algorithm's performance on the equation~\ref{eq:bessel_composite} composite Bessel function regression varies with learning rate $\eta$, training epochs $E$, and number of hidden neurons $N$. The top row displays the difference $\delta \loss$ between the mean final losses of growing and static networks, with blue regions indicating where growing networks outperform static ones. The bottom row tracks the value from the controller $C_1$ over the training epochs. 

When using adaptive optimizers, the learning rate $\eta$ emerges as a critical factor in the growth dynamics. With small learning rates, $C_1$ increases slowly, keeping the network small throughout the training period. Large learning rates on the other hand cause rapid growth, quickly matching the static network's size and hence yielding a similar performance. The optimal regime occurs in the intermediate range, where growing networks reach full size towards the end of the training and outperform static networks. The transition threshold appears consistently at approximately $C_1\approx 0.7$ for this problem, as highlighted by the dashed lines.

This learning rate dependence just as the size-loss coupling dependence before demonstrates that the timing of the network growth relative to the training process is a crucial factor for performance optimization.

\subsection{Network Size Relations}
To demonstrate the fundamental relationships between network size, network performance, and task complexity, we systematically vary the number of hidden neurons $N$ and the number of classification classes~(spirals) $N_c$ to find a surprising relationship between the accuracy of growing to static networks in the spiral classification problem.

 \begin{figure*}[!htb]
    \centering
    \includegraphics[width=1.\linewidth]{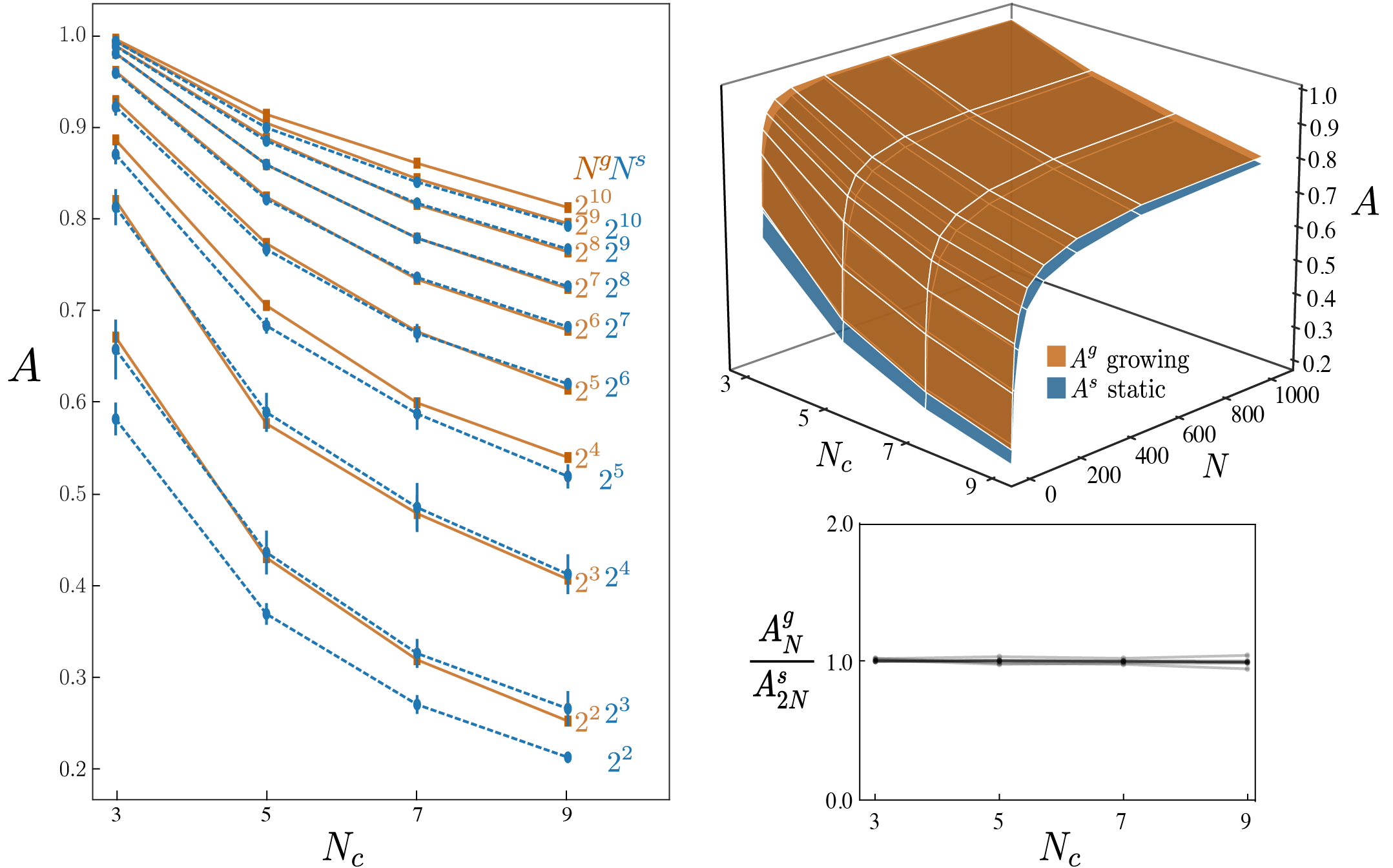}
   \vspace{-0.5cm}
    \caption{Scaling of controller-mask spiral pattern classification. 3D plot shows that accuracy $A$ increases with increasing neurons $N$ and decreases with increasing number of classes $N_c$ but with growing outperforming static always. 2D plots show cross sections and accuracy ratio. Growing is as accurate as static with just half the neurons.}
    \label{fig:controllerAlgSpiralScaling}
\end{figure*}

Figure~\ref{fig:controllerAlgSpiralScaling} demonstrates the relationship between network size and performance in growing and static networks across spiral classification task complexity. The 3D visualization reveals that the classification accuracy $A$ increases systematically with neuron count $N$ while decreasing with the number of classes $N_c$.
The 2D cross-sectional and ratio plots establish a remarkable efficiency advantage: Growing networks achieve comparable accuracy to static networks while using only half the neurons, following the relationship
\begin{equation} 
\label{eq:size_efficiency} 
A^g_N \approx A^s_{2N}
\end{equation}
for this classification task.

\clearpage

\section{Discussion}
\label{sec:dis}

Our investigation of growing neural networks reveals several fundamental insights about neural network optimization. First, the superior performance of growing networks over static networks of equivalent final size suggests that the trajectory of learning may be as important as the final architecture. Starting small and growing naturally appears to help avoid many of the local minima that plague larger static networks during training.

The effectiveness of both our auxiliary-weight and controller-mask implementations demonstrates that continuous, gradient-based size optimization is not only possible but potentially preferable to discrete architectural choices. This challenges the common practice of selecting network sizes through trial and error or extensive architecture searches.

The memory and computational efficiency gains observed in our growing networks have significant practical implications. In an era where AI models are growing exponentially in size and energy consumption, our results suggest that ``bigger is better'' may not be the only path forward. Growing networks offer a way to automatically find efficient architectures that balance complexity with performance.

However, the choice between auxiliary-weight and controller-mask implementations presents important trade-offs. The auxiliary-weight approach offers better memory efficiency for small networks and more interpretable growth dynamics, while the controller-mask implementation provides better stability and scaling to larger problems. These trade-offs must be considered in the context of specific application requirements.

Our work spans several domains. The smooth transition functions we employ connect to recent work on continuous relaxations in neural architecture search and AutoML more generally. The gradient-based size optimization relates to meta-learning and hyperparameter optimization. The superior performance of growing networks may offer insights into the loss landscape geometry of neural networks and how it changes with network size.

Several limitations of our current approach warrant discussion. First, while our methods work well for feed-forward networks, extending them to more complex architectures like convolutional or recurrent networks presents additional challenges. Second, the choice of size-dependent loss terms and their coupling strengths can impact performance, and optimal selection of these parameters remains an open question.

This work opens several promising research directions. First, our growing network framework could be extended to other architectural parameters beyond just network width. Methods for layer-wise size optimization and bidirectional size adjustment (both growth and pruning) could provide even more flexible architecture evolution. The relationship between network growth and optimization dynamics also warrants deeper theoretical investigation, particularly regarding expressivity guarantees and convergence properties.

Recent work has revealed that curriculum learning shows diminishing returns as model size increases, with large models able to effectively memorize both easy and difficult examples simultaneously~\cite{saraomannelli2024tilting}. Growing networks may offer a natural solution to this challenge. The initial small network size could enforce architectural constraints that align with curriculum progression, preventing premature memorization of complex patterns. As the network grows, its increasing capacity would parallel the curriculum's progression toward more difficult examples. This potential synergy between architectural growth and curriculum learning raises fundamental questions about the relationship between network capacity and learning dynamics.

The practical applications of growing networks extend beyond theoretical interest. Scaling these approaches to larger problems requires careful consideration of implementation efficiency and hardware acceleration. The development of specialized hardware architectures that can efficiently handle dynamic network sizes could significantly impact deployment scenarios. Integration with other efficiency-focused techniques, such as quantization and pruning, could further enhance the practical utility of growing networks.

Finally, as AI systems continue to grow in size and energy consumption, the sustainability implications of our approach deserve investigation~\cite{varoquaux2024hype}. Quantitative analysis of energy efficiency gains, particularly in edge computing scenarios where resource constraints are significant, could help establish growing networks as a practical tool for sustainable AI development. The development of comprehensive metrics for architectural efficiency would allow better comparison between different approaches to network size optimization.

\begin{framed}
\noindent After completing this manuscript, related work~\cite{evci2022GradMax} came to our attention.
\end{framed}

\begin{acknowledgments}
This work was funded in part by a gift from United Therapeutics. We thank Matthew Gill for fruitful discussions.
\end{acknowledgments}

\vspace{1cm}
\appendix
\section{Size influence parameter invariance in adaptive optimizers }
\label{app:adaptive}
When using adaptive optimizers like Adam, the size influence parameter $\lambda$ becomes effectively redundant. This occurs because adaptive optimizers automatically scale gradients based on their running moments. For Adam, with first moment $m \sim \nabla$, second moment $v \sim \nabla^2$, and descent $\theta \leftarrow \theta-\eta\  m/\sqrt{v}$, the effective gradient for size-influence-$\lambda$ loss terms is
\begin{equation}
    \frac{m}{\sqrt{v}} \sim \frac{\lambda\nabla}{\sqrt{\lambda^2\nabla^2}} = \text{sign}(\nabla),
\end{equation}
where $\nabla$ represents the raw gradient. The $\lambda$ terms cancel out in the ratio, making the optimization process invariant to the choice of $\lambda$. This means that when using adaptive optimizers, $\lambda$ can be set to $1$ without loss of generality, and the relative scale of size and task gradients is automatically balanced.

\bibliography{GrowingNetworks} % Produces the bibliography via BibTeX.

%merlin.mbs apsrev4-1.bst 2010-07-25 4.21a (PWD, AO, DPC) hacked
%Control: key (0)
%Control: author (8) initials jnrlst
%Control: editor formatted (1) identically to author
%Control: production of article title (-1) disabled
%Control: page (0) single
%Control: year (1) truncated
%Control: production of eprint (0) enabled
\begin{thebibliography}{28}%
\makeatletter
\providecommand \@ifxundefined [1]{%
 \@ifx{#1\undefined}
}%
\providecommand \@ifnum [1]{%
 \ifnum #1\expandafter \@firstoftwo
 \else \expandafter \@secondoftwo
 \fi
}%
\providecommand \@ifx [1]{%
 \ifx #1\expandafter \@firstoftwo
 \else \expandafter \@secondoftwo
 \fi
}%
\providecommand \natexlab [1]{#1}%
\providecommand \enquote  [1]{``#1''}%
\providecommand \bibnamefont  [1]{#1}%
\providecommand \bibfnamefont [1]{#1}%
\providecommand \citenamefont [1]{#1}%
\providecommand \href@noop [0]{\@secondoftwo}%
\providecommand \href [0]{\begingroup \@sanitize@url \@href}%
\providecommand \@href[1]{\@@startlink{#1}\@@href}%
\providecommand \@@href[1]{\endgroup#1\@@endlink}%
\providecommand \@sanitize@url [0]{\catcode `\\12\catcode `\$12\catcode
  `\&12\catcode `\#12\catcode `\^12\catcode `\_12\catcode `\%12\relax}%
\providecommand \@@startlink[1]{}%
\providecommand \@@endlink[0]{}%
\providecommand \url  [0]{\begingroup\@sanitize@url \@url }%
\providecommand \@url [1]{\endgroup\@href {#1}{\urlprefix }}%
\providecommand \urlprefix  [0]{URL }%
\providecommand \Eprint [0]{\href }%
\providecommand \doibase [0]{http://dx.doi.org/}%
\providecommand \selectlanguage [0]{\@gobble}%
\providecommand \bibinfo  [0]{\@secondoftwo}%
\providecommand \bibfield  [0]{\@secondoftwo}%
\providecommand \translation [1]{[#1]}%
\providecommand \BibitemOpen [0]{}%
\providecommand \bibitemStop [0]{}%
\providecommand \bibitemNoStop [0]{.\EOS\space}%
\providecommand \EOS [0]{\spacefactor3000\relax}%
\providecommand \BibitemShut  [1]{\csname bibitem#1\endcsname}%
\let\auto@bib@innerbib\@empty
%</preamble>
\bibitem [{\citenamefont {Brown}\ \emph {et~al.}(2020)\citenamefont {Brown},
  \citenamefont {Mann}, \citenamefont {Ryder}, \citenamefont {Subbiah},
  \citenamefont {Kaplan}, \citenamefont {Dhariwal}, \citenamefont
  {Neelakantan}, \citenamefont {Shyam}, \citenamefont {Sastry}, \citenamefont
  {Askell}, \citenamefont {Agarwal}, \citenamefont {Herbert-Voss},
  \citenamefont {Krueger}, \citenamefont {Henighan}, \citenamefont {Child},
  \citenamefont {Ramesh}, \citenamefont {Ziegler}, \citenamefont {Wu},
  \citenamefont {Winter}, \citenamefont {Hesse}, \citenamefont {Chen},
  \citenamefont {Sigler}, \citenamefont {Litwin}, \citenamefont {Gray},
  \citenamefont {Chess}, \citenamefont {Clark}, \citenamefont {Berner},
  \citenamefont {McCandlish}, \citenamefont {Radford}, \citenamefont
  {Sutskever},\ and\ \citenamefont {Amodei}}]{brown2020language}%
  \BibitemOpen
  \bibfield  {author} {\bibinfo {author} {\bibfnamefont {T.~B.}\ \bibnamefont
  {Brown}}, \bibinfo {author} {\bibfnamefont {B.}~\bibnamefont {Mann}},
  \bibinfo {author} {\bibfnamefont {N.}~\bibnamefont {Ryder}}, \bibinfo
  {author} {\bibfnamefont {M.}~\bibnamefont {Subbiah}}, \bibinfo {author}
  {\bibfnamefont {J.}~\bibnamefont {Kaplan}}, \bibinfo {author} {\bibfnamefont
  {P.}~\bibnamefont {Dhariwal}}, \bibinfo {author} {\bibfnamefont
  {A.}~\bibnamefont {Neelakantan}}, \bibinfo {author} {\bibfnamefont
  {P.}~\bibnamefont {Shyam}}, \bibinfo {author} {\bibfnamefont
  {G.}~\bibnamefont {Sastry}}, \bibinfo {author} {\bibfnamefont
  {A.}~\bibnamefont {Askell}}, \bibinfo {author} {\bibfnamefont
  {S.}~\bibnamefont {Agarwal}}, \bibinfo {author} {\bibfnamefont
  {A.}~\bibnamefont {Herbert-Voss}}, \bibinfo {author} {\bibfnamefont
  {G.}~\bibnamefont {Krueger}}, \bibinfo {author} {\bibfnamefont
  {T.}~\bibnamefont {Henighan}}, \bibinfo {author} {\bibfnamefont
  {R.}~\bibnamefont {Child}}, \bibinfo {author} {\bibfnamefont
  {A.}~\bibnamefont {Ramesh}}, \bibinfo {author} {\bibfnamefont {D.~M.}\
  \bibnamefont {Ziegler}}, \bibinfo {author} {\bibfnamefont {J.}~\bibnamefont
  {Wu}}, \bibinfo {author} {\bibfnamefont {C.}~\bibnamefont {Winter}}, \bibinfo
  {author} {\bibfnamefont {C.}~\bibnamefont {Hesse}}, \bibinfo {author}
  {\bibfnamefont {M.}~\bibnamefont {Chen}}, \bibinfo {author} {\bibfnamefont
  {E.}~\bibnamefont {Sigler}}, \bibinfo {author} {\bibfnamefont
  {M.}~\bibnamefont {Litwin}}, \bibinfo {author} {\bibfnamefont
  {S.}~\bibnamefont {Gray}}, \bibinfo {author} {\bibfnamefont {B.}~\bibnamefont
  {Chess}}, \bibinfo {author} {\bibfnamefont {J.}~\bibnamefont {Clark}},
  \bibinfo {author} {\bibfnamefont {C.}~\bibnamefont {Berner}}, \bibinfo
  {author} {\bibfnamefont {S.}~\bibnamefont {McCandlish}}, \bibinfo {author}
  {\bibfnamefont {A.}~\bibnamefont {Radford}}, \bibinfo {author} {\bibfnamefont
  {I.}~\bibnamefont {Sutskever}}, \ and\ \bibinfo {author} {\bibfnamefont
  {D.}~\bibnamefont {Amodei}},\ }\href@noop {} {\bibfield  {journal} {\bibinfo
  {journal} {arXiv:2005.14165}\ } (\bibinfo {year} {2020})}\BibitemShut
  {NoStop}%
\bibitem [{\citenamefont {Strubell}\ \emph {et~al.}(2019)\citenamefont
  {Strubell}, \citenamefont {Ganesh},\ and\ \citenamefont
  {McCallum}}]{strubell2019energy}%
  \BibitemOpen
  \bibfield  {author} {\bibinfo {author} {\bibfnamefont {E.}~\bibnamefont
  {Strubell}}, \bibinfo {author} {\bibfnamefont {A.}~\bibnamefont {Ganesh}}, \
  and\ \bibinfo {author} {\bibfnamefont {A.}~\bibnamefont {McCallum}},\
  }\href@noop {} {\bibfield  {journal} {\bibinfo  {journal} {arXiv:1906.02243}\
  } (\bibinfo {year} {2019})}\BibitemShut {NoStop}%
\bibitem [{\citenamefont {Amodei}\ and\ \citenamefont
  {Hernandez}(2018)}]{openai2018ai}%
  \BibitemOpen
  \bibfield  {author} {\bibinfo {author} {\bibfnamefont {D.}~\bibnamefont
  {Amodei}}\ and\ \bibinfo {author} {\bibfnamefont {D.}~\bibnamefont
  {Hernandez}},\ }\href {https://openai.com/index/ai-and-compute/#modern}
  {\enquote {\bibinfo {title} {Ai and compute},}\ } (\bibinfo {year} {2018}),\
  \bibinfo {note} {accessed: 2025-02-14}\BibitemShut {NoStop}%
\bibitem [{\citenamefont {Frankle}\ and\ \citenamefont
  {Carbin}(2018)}]{frankle2018lottery}%
  \BibitemOpen
  \bibfield  {author} {\bibinfo {author} {\bibfnamefont {J.}~\bibnamefont
  {Frankle}}\ and\ \bibinfo {author} {\bibfnamefont {M.}~\bibnamefont
  {Carbin}},\ }\href@noop {} {\bibfield  {journal} {\bibinfo  {journal}
  {arXiv:1803.03635}\ } (\bibinfo {year} {2018})}\BibitemShut {NoStop}%
\bibitem [{\citenamefont {See}\ \emph {et~al.}(2016)\citenamefont {See},
  \citenamefont {Luong},\ and\ \citenamefont {Manning}}]{see2016compression}%
  \BibitemOpen
  \bibfield  {author} {\bibinfo {author} {\bibfnamefont {A.}~\bibnamefont
  {See}}, \bibinfo {author} {\bibfnamefont {M.-T.}\ \bibnamefont {Luong}}, \
  and\ \bibinfo {author} {\bibfnamefont {C.~D.}\ \bibnamefont {Manning}},\
  }\href@noop {} {\bibfield  {journal} {\bibinfo  {journal} {arXiv:1606.09274}\
  } (\bibinfo {year} {2016})}\BibitemShut {NoStop}%
\bibitem [{\citenamefont {Zhao}\ \emph {et~al.}(2024)\citenamefont {Zhao},
  \citenamefont {Wang}, \citenamefont {Jiang}, \citenamefont {Ma},
  \citenamefont {Ma}, \citenamefont {He}, \citenamefont {Du}, \citenamefont
  {Ma},\ and\ \citenamefont {Huang}}]{zhao2024integrative}%
  \BibitemOpen
  \bibfield  {author} {\bibinfo {author} {\bibfnamefont {M.}~\bibnamefont
  {Zhao}}, \bibinfo {author} {\bibfnamefont {N.}~\bibnamefont {Wang}}, \bibinfo
  {author} {\bibfnamefont {X.}~\bibnamefont {Jiang}}, \bibinfo {author}
  {\bibfnamefont {X.}~\bibnamefont {Ma}}, \bibinfo {author} {\bibfnamefont
  {H.}~\bibnamefont {Ma}}, \bibinfo {author} {\bibfnamefont {G.}~\bibnamefont
  {He}}, \bibinfo {author} {\bibfnamefont {K.}~\bibnamefont {Du}}, \bibinfo
  {author} {\bibfnamefont {L.}~\bibnamefont {Ma}}, \ and\ \bibinfo {author}
  {\bibfnamefont {T.}~\bibnamefont {Huang}},\ }\href {\doibase
  10.1038/s43588-024-00738-w} {\bibfield  {journal} {\bibinfo  {journal}
  {Nature Computational Science}\ }\textbf {\bibinfo {volume} {4}},\ \bibinfo
  {pages} {978–990} (\bibinfo {year} {2024})}\BibitemShut {NoStop}%
\bibitem [{\citenamefont
  {Gebicke-Haerter}(2023)}]{gebickehaerter2023computational}%
  \BibitemOpen
  \bibfield  {author} {\bibinfo {author} {\bibfnamefont {P.~J.}\ \bibnamefont
  {Gebicke-Haerter}},\ }\href {\doibase 10.3389/fncel.2023.1220030} {\bibfield
  {journal} {\bibinfo  {journal} {Frontiers in Cellular Neuroscience}\ }\textbf
  {\bibinfo {volume} {17}} (\bibinfo {year} {2023}),\
  10.3389/fncel.2023.1220030}\BibitemShut {NoStop}%
\bibitem [{\citenamefont {Fan}\ \emph {et~al.}(2004)\citenamefont {Fan},
  \citenamefont {Chen},\ and\ \citenamefont {Ko}}]{fan2004evolving}%
  \BibitemOpen
  \bibfield  {author} {\bibinfo {author} {\bibfnamefont {Z.}~\bibnamefont
  {Fan}}, \bibinfo {author} {\bibfnamefont {G.}~\bibnamefont {Chen}}, \ and\
  \bibinfo {author} {\bibfnamefont {K.~T.}\ \bibnamefont {Ko}},\ }\href
  {\doibase 10.1007/s11768-004-0024-8} {\bibfield  {journal} {\bibinfo
  {journal} {Journal of Control Theory and Applications}\ }\textbf {\bibinfo
  {volume} {2}},\ \bibinfo {pages} {60–64} (\bibinfo {year}
  {2004})}\BibitemShut {NoStop}%
\bibitem [{\citenamefont {Aggarwal}\ and\ \citenamefont
  {Subbian}(2014)}]{aggarwal2014evolutionary}%
  \BibitemOpen
  \bibfield  {author} {\bibinfo {author} {\bibfnamefont {C.}~\bibnamefont
  {Aggarwal}}\ and\ \bibinfo {author} {\bibfnamefont {K.}~\bibnamefont
  {Subbian}},\ }\href {\doibase 10.1145/2601412} {\bibfield  {journal}
  {\bibinfo  {journal} {ACM Comput. Surv.}\ }\textbf {\bibinfo {volume} {47}}
  (\bibinfo {year} {2014}),\ 10.1145/2601412}\BibitemShut {NoStop}%
\bibitem [{\citenamefont {Tunc}\ and\ \citenamefont
  {Shaw}(2014)}]{tunc2014effects}%
  \BibitemOpen
  \bibfield  {author} {\bibinfo {author} {\bibfnamefont {I.}~\bibnamefont
  {Tunc}}\ and\ \bibinfo {author} {\bibfnamefont {L.~B.}\ \bibnamefont
  {Shaw}},\ }\href {\doibase 10.1103/physreve.90.022801} {\bibfield  {journal}
  {\bibinfo  {journal} {Physical Review E}\ }\textbf {\bibinfo {volume} {90}}
  (\bibinfo {year} {2014}),\ 10.1103/physreve.90.022801}\BibitemShut {NoStop}%
\bibitem [{\citenamefont {Elsken}\ \emph {et~al.}(2018)\citenamefont {Elsken},
  \citenamefont {Metzen},\ and\ \citenamefont {Hutter}}]{elsken2018neural}%
  \BibitemOpen
  \bibfield  {author} {\bibinfo {author} {\bibfnamefont {T.}~\bibnamefont
  {Elsken}}, \bibinfo {author} {\bibfnamefont {J.~H.}\ \bibnamefont {Metzen}},
  \ and\ \bibinfo {author} {\bibfnamefont {F.}~\bibnamefont {Hutter}},\
  }\href@noop {} {\bibfield  {journal} {\bibinfo  {journal} {arXiv:1808.05377}\
  } (\bibinfo {year} {2018})}\BibitemShut {NoStop}%
\bibitem [{\citenamefont {Blalock}\ \emph {et~al.}(2020)\citenamefont
  {Blalock}, \citenamefont {Ortiz}, \citenamefont {Frankle},\ and\
  \citenamefont {Guttag}}]{blalock2020what}%
  \BibitemOpen
  \bibfield  {author} {\bibinfo {author} {\bibfnamefont {D.}~\bibnamefont
  {Blalock}}, \bibinfo {author} {\bibfnamefont {J.~J.~G.}\ \bibnamefont
  {Ortiz}}, \bibinfo {author} {\bibfnamefont {J.}~\bibnamefont {Frankle}}, \
  and\ \bibinfo {author} {\bibfnamefont {J.}~\bibnamefont {Guttag}},\
  }\href@noop {} {\bibfield  {journal} {\bibinfo  {journal} {arXiv:2003.03033}\
  } (\bibinfo {year} {2020})}\BibitemShut {NoStop}%
\bibitem [{\citenamefont {Stanley}\ and\ \citenamefont
  {Miikkulainen}(2002)}]{stanley2002evolving}%
  \BibitemOpen
  \bibfield  {author} {\bibinfo {author} {\bibfnamefont {K.~O.}\ \bibnamefont
  {Stanley}}\ and\ \bibinfo {author} {\bibfnamefont {R.}~\bibnamefont
  {Miikkulainen}},\ }\href {\doibase 10.1162/106365602320169811} {\bibfield
  {journal} {\bibinfo  {journal} {Evolutionary Computation}\ }\textbf {\bibinfo
  {volume} {10}},\ \bibinfo {pages} {99–127} (\bibinfo {year}
  {2002})}\BibitemShut {NoStop}%
\bibitem [{\citenamefont {Fahlman}\ and\ \citenamefont
  {Lebiere}(1989)}]{fahlman1989cascade}%
  \BibitemOpen
  \bibfield  {author} {\bibinfo {author} {\bibfnamefont {S.}~\bibnamefont
  {Fahlman}}\ and\ \bibinfo {author} {\bibfnamefont {C.}~\bibnamefont
  {Lebiere}},\ }in\ \href
  {https://proceedings.neurips.cc/paper_files/paper/1989/file/69adc1e107f7f7d035d7baf04342e1ca-Paper.pdf}
  {\emph {\bibinfo {booktitle} {Advances in Neural Information Processing
  Systems}}},\ Vol.~\bibinfo {volume} {2},\ \bibinfo {editor} {edited by\
  \bibinfo {editor} {\bibfnamefont {D.}~\bibnamefont {Touretzky}}}\ (\bibinfo
  {publisher} {Morgan-Kaufmann},\ \bibinfo {year} {1989})\BibitemShut {NoStop}%
\bibitem [{\citenamefont {Bottou}\ \emph {et~al.}(2016)\citenamefont {Bottou},
  \citenamefont {Curtis},\ and\ \citenamefont
  {Nocedal}}]{bottou2016optimization}%
  \BibitemOpen
  \bibfield  {author} {\bibinfo {author} {\bibfnamefont {L.}~\bibnamefont
  {Bottou}}, \bibinfo {author} {\bibfnamefont {F.~E.}\ \bibnamefont {Curtis}},
  \ and\ \bibinfo {author} {\bibfnamefont {J.}~\bibnamefont {Nocedal}},\
  }\href@noop {} {\bibfield  {journal} {\bibinfo  {journal} {arXiv:1606.04838}\
  } (\bibinfo {year} {2016})}\BibitemShut {NoStop}%
\bibitem [{\citenamefont {Allen-Zhu}\ \emph {et~al.}(2019)\citenamefont
  {Allen-Zhu}, \citenamefont {Li},\ and\ \citenamefont
  {Song}}]{allen2019convergence}%
  \BibitemOpen
  \bibfield  {author} {\bibinfo {author} {\bibfnamefont {Z.}~\bibnamefont
  {Allen-Zhu}}, \bibinfo {author} {\bibfnamefont {Y.}~\bibnamefont {Li}}, \
  and\ \bibinfo {author} {\bibfnamefont {Z.}~\bibnamefont {Song}},\ }in\ \href
  {https://proceedings.mlr.press/v97/allen-zhu19a.html} {\emph {\bibinfo
  {booktitle} {Proceedings of the 36th International Conference on Machine
  Learning}}},\ \bibinfo {series} {Proceedings of Machine Learning Research},
  Vol.~\bibinfo {volume} {97},\ \bibinfo {editor} {edited by\ \bibinfo {editor}
  {\bibfnamefont {K.}~\bibnamefont {Chaudhuri}}\ and\ \bibinfo {editor}
  {\bibfnamefont {R.}~\bibnamefont {Salakhutdinov}}}\ (\bibinfo  {publisher}
  {PMLR},\ \bibinfo {year} {2019})\ pp.\ \bibinfo {pages}
  {242--252}\BibitemShut {NoStop}%
\bibitem [{\citenamefont {Ghadimi}\ and\ \citenamefont
  {Lan}(2013)}]{ghadimi2013stochastic}%
  \BibitemOpen
  \bibfield  {author} {\bibinfo {author} {\bibfnamefont {S.}~\bibnamefont
  {Ghadimi}}\ and\ \bibinfo {author} {\bibfnamefont {G.}~\bibnamefont {Lan}},\
  }\href@noop {} {\bibfield  {journal} {\bibinfo  {journal} {arXiv:1309.5549}\
  } (\bibinfo {year} {2013})}\BibitemShut {NoStop}%
\bibitem [{\citenamefont {E}\ \emph {et~al.}(2019)\citenamefont {E},
  \citenamefont {Ma},\ and\ \citenamefont {Wu}}]{e2019comparative}%
  \BibitemOpen
  \bibfield  {author} {\bibinfo {author} {\bibfnamefont {W.}~\bibnamefont {E}},
  \bibinfo {author} {\bibfnamefont {C.}~\bibnamefont {Ma}}, \ and\ \bibinfo
  {author} {\bibfnamefont {L.}~\bibnamefont {Wu}},\ }\href@noop {} {\bibfield
  {journal} {\bibinfo  {journal} {arXiv:1904.04326}\ } (\bibinfo {year}
  {2019})}\BibitemShut {NoStop}%
\bibitem [{\citenamefont {Jin}\ \emph {et~al.}(2022)\citenamefont {Jin},
  \citenamefont {Mattheakis},\ and\ \citenamefont
  {Protopapas}}]{jin2022physicsinformed}%
  \BibitemOpen
  \bibfield  {author} {\bibinfo {author} {\bibfnamefont {H.}~\bibnamefont
  {Jin}}, \bibinfo {author} {\bibfnamefont {M.}~\bibnamefont {Mattheakis}}, \
  and\ \bibinfo {author} {\bibfnamefont {P.}~\bibnamefont {Protopapas}},\
  }\href@noop {} {\bibfield  {journal} {\bibinfo  {journal} {arXiv:2203.00451}\
  } (\bibinfo {year} {2022})}\BibitemShut {NoStop}%
\bibitem [{Mat()}]{Mathematica}%
  \BibitemOpen
  \href {https://www.wolfram.com/mathematica} {\enquote {\bibinfo {title}
  {{W}olfram {R}esearch, {I}nc.{,} {M}athematica},}\ }\bibinfo {note}
  {{C}hampaign, IL, 2024}\BibitemShut {NoStop}%
\bibitem [{\citenamefont {Bradbury}\ \emph {et~al.}(2018)\citenamefont
  {Bradbury}, \citenamefont {Frostig}, \citenamefont {Hawkins}, \citenamefont
  {Johnson}, \citenamefont {Leary}, \citenamefont {Maclaurin}, \citenamefont
  {Necula}, \citenamefont {Paszke}, \citenamefont {Vander{P}las}, \citenamefont
  {Wanderman-{M}ilne},\ and\ \citenamefont {Zhang}}]{jax2018github}%
  \BibitemOpen
  \bibfield  {author} {\bibinfo {author} {\bibfnamefont {J.}~\bibnamefont
  {Bradbury}}, \bibinfo {author} {\bibfnamefont {R.}~\bibnamefont {Frostig}},
  \bibinfo {author} {\bibfnamefont {P.}~\bibnamefont {Hawkins}}, \bibinfo
  {author} {\bibfnamefont {M.~J.}\ \bibnamefont {Johnson}}, \bibinfo {author}
  {\bibfnamefont {C.}~\bibnamefont {Leary}}, \bibinfo {author} {\bibfnamefont
  {D.}~\bibnamefont {Maclaurin}}, \bibinfo {author} {\bibfnamefont
  {G.}~\bibnamefont {Necula}}, \bibinfo {author} {\bibfnamefont
  {A.}~\bibnamefont {Paszke}}, \bibinfo {author} {\bibfnamefont
  {J.}~\bibnamefont {Vander{P}las}}, \bibinfo {author} {\bibfnamefont
  {S.}~\bibnamefont {Wanderman-{M}ilne}}, \ and\ \bibinfo {author}
  {\bibfnamefont {Q.}~\bibnamefont {Zhang}},\ }\href
  {http://github.com/google/jax} {\enquote {\bibinfo {title} {{JAX}: composable
  transformations of {P}ython+{N}um{P}y programs},}\ } (\bibinfo {year}
  {2018})\BibitemShut {NoStop}%
\bibitem [{\citenamefont {Kidger}\ and\ \citenamefont
  {Garcia}(2021)}]{kidger2021equinox}%
  \BibitemOpen
  \bibfield  {author} {\bibinfo {author} {\bibfnamefont {P.}~\bibnamefont
  {Kidger}}\ and\ \bibinfo {author} {\bibfnamefont {C.}~\bibnamefont
  {Garcia}},\ }\href@noop {} {\bibfield  {journal} {\bibinfo  {journal}
  {Differentiable Programming workshop at Neural Information Processing Systems
  2021}\ } (\bibinfo {year} {2021})}\BibitemShut {NoStop}%
\bibitem [{\citenamefont {DeepMind}\ \emph {et~al.}(2020)\citenamefont
  {DeepMind}, \citenamefont {Babuschkin}, \citenamefont {Baumli}, \citenamefont
  {Bell}, \citenamefont {Bhupatiraju}, \citenamefont {Bruce}, \citenamefont
  {Buchlovsky}, \citenamefont {Budden}, \citenamefont {Cai}, \citenamefont
  {Clark}, \citenamefont {Danihelka}, \citenamefont {Dedieu}, \citenamefont
  {Fantacci}, \citenamefont {Godwin}, \citenamefont {Jones}, \citenamefont
  {Hemsley}, \citenamefont {Hennigan}, \citenamefont {Hessel}, \citenamefont
  {Hou}, \citenamefont {Kapturowski}, \citenamefont {Keck}, \citenamefont
  {Kemaev}, \citenamefont {King}, \citenamefont {Kunesch}, \citenamefont
  {Martens}, \citenamefont {Merzic}, \citenamefont {Mikulik}, \citenamefont
  {Norman}, \citenamefont {Papamakarios}, \citenamefont {Quan}, \citenamefont
  {Ring}, \citenamefont {Ruiz}, \citenamefont {Sanchez}, \citenamefont
  {Sartran}, \citenamefont {Schneider}, \citenamefont {Sezener}, \citenamefont
  {Spencer}, \citenamefont {Srinivasan}, \citenamefont {Stanojevi\'{c}},
  \citenamefont {Stokowiec}, \citenamefont {Wang}, \citenamefont {Zhou},\ and\
  \citenamefont {Viola}}]{deepmind2020jax}%
  \BibitemOpen
  \bibfield  {author} {\bibinfo {author} {\bibnamefont {DeepMind}}, \bibinfo
  {author} {\bibfnamefont {I.}~\bibnamefont {Babuschkin}}, \bibinfo {author}
  {\bibfnamefont {K.}~\bibnamefont {Baumli}}, \bibinfo {author} {\bibfnamefont
  {A.}~\bibnamefont {Bell}}, \bibinfo {author} {\bibfnamefont {S.}~\bibnamefont
  {Bhupatiraju}}, \bibinfo {author} {\bibfnamefont {J.}~\bibnamefont {Bruce}},
  \bibinfo {author} {\bibfnamefont {P.}~\bibnamefont {Buchlovsky}}, \bibinfo
  {author} {\bibfnamefont {D.}~\bibnamefont {Budden}}, \bibinfo {author}
  {\bibfnamefont {T.}~\bibnamefont {Cai}}, \bibinfo {author} {\bibfnamefont
  {A.}~\bibnamefont {Clark}}, \bibinfo {author} {\bibfnamefont
  {I.}~\bibnamefont {Danihelka}}, \bibinfo {author} {\bibfnamefont
  {A.}~\bibnamefont {Dedieu}}, \bibinfo {author} {\bibfnamefont
  {C.}~\bibnamefont {Fantacci}}, \bibinfo {author} {\bibfnamefont
  {J.}~\bibnamefont {Godwin}}, \bibinfo {author} {\bibfnamefont
  {C.}~\bibnamefont {Jones}}, \bibinfo {author} {\bibfnamefont
  {R.}~\bibnamefont {Hemsley}}, \bibinfo {author} {\bibfnamefont
  {T.}~\bibnamefont {Hennigan}}, \bibinfo {author} {\bibfnamefont
  {M.}~\bibnamefont {Hessel}}, \bibinfo {author} {\bibfnamefont
  {S.}~\bibnamefont {Hou}}, \bibinfo {author} {\bibfnamefont {S.}~\bibnamefont
  {Kapturowski}}, \bibinfo {author} {\bibfnamefont {T.}~\bibnamefont {Keck}},
  \bibinfo {author} {\bibfnamefont {I.}~\bibnamefont {Kemaev}}, \bibinfo
  {author} {\bibfnamefont {M.}~\bibnamefont {King}}, \bibinfo {author}
  {\bibfnamefont {M.}~\bibnamefont {Kunesch}}, \bibinfo {author} {\bibfnamefont
  {L.}~\bibnamefont {Martens}}, \bibinfo {author} {\bibfnamefont
  {H.}~\bibnamefont {Merzic}}, \bibinfo {author} {\bibfnamefont
  {V.}~\bibnamefont {Mikulik}}, \bibinfo {author} {\bibfnamefont
  {T.}~\bibnamefont {Norman}}, \bibinfo {author} {\bibfnamefont
  {G.}~\bibnamefont {Papamakarios}}, \bibinfo {author} {\bibfnamefont
  {J.}~\bibnamefont {Quan}}, \bibinfo {author} {\bibfnamefont {R.}~\bibnamefont
  {Ring}}, \bibinfo {author} {\bibfnamefont {F.}~\bibnamefont {Ruiz}}, \bibinfo
  {author} {\bibfnamefont {A.}~\bibnamefont {Sanchez}}, \bibinfo {author}
  {\bibfnamefont {L.}~\bibnamefont {Sartran}}, \bibinfo {author} {\bibfnamefont
  {R.}~\bibnamefont {Schneider}}, \bibinfo {author} {\bibfnamefont
  {E.}~\bibnamefont {Sezener}}, \bibinfo {author} {\bibfnamefont
  {S.}~\bibnamefont {Spencer}}, \bibinfo {author} {\bibfnamefont
  {S.}~\bibnamefont {Srinivasan}}, \bibinfo {author} {\bibfnamefont
  {M.}~\bibnamefont {Stanojevi\'{c}}}, \bibinfo {author} {\bibfnamefont
  {W.}~\bibnamefont {Stokowiec}}, \bibinfo {author} {\bibfnamefont
  {L.}~\bibnamefont {Wang}}, \bibinfo {author} {\bibfnamefont {G.}~\bibnamefont
  {Zhou}}, \ and\ \bibinfo {author} {\bibfnamefont {F.}~\bibnamefont {Viola}},\
  }\href {http://github.com/google-deepmind} {\enquote {\bibinfo {title} {The
  {D}eep{M}ind {JAX} {E}cosystem},}\ } (\bibinfo {year} {2020})\BibitemShut
  {NoStop}%
\bibitem [{Note1()}]{Note1}%
  \BibitemOpen
  \bibinfo {note} {Our code is available at our GitHub repository
  https://github.com/NonlinearArtificialIntelligenceLab/N3}\BibitemShut
  {NoStop}%
\bibitem [{\citenamefont {Chalup}\ and\ \citenamefont
  {Wiklendt}(2007)}]{chalup2007variations}%
  \BibitemOpen
  \bibfield  {author} {\bibinfo {author} {\bibfnamefont {S.~K.}\ \bibnamefont
  {Chalup}}\ and\ \bibinfo {author} {\bibfnamefont {L.}~\bibnamefont
  {Wiklendt}},\ }\href {\doibase 10.1080/09540090701398017} {\bibfield
  {journal} {\bibinfo  {journal} {Connection Science}\ }\textbf {\bibinfo
  {volume} {19}},\ \bibinfo {pages} {183–199} (\bibinfo {year}
  {2007})}\BibitemShut {NoStop}%
\bibitem [{\citenamefont {Sarao~Mannelli}\ \emph {et~al.}(2024)\citenamefont
  {Sarao~Mannelli}, \citenamefont {Ivashynka}, \citenamefont {Saxe},\ and\
  \citenamefont {Saglietti}}]{saraomannelli2024tilting}%
  \BibitemOpen
  \bibfield  {author} {\bibinfo {author} {\bibfnamefont {S.}~\bibnamefont
  {Sarao~Mannelli}}, \bibinfo {author} {\bibfnamefont {Y.}~\bibnamefont
  {Ivashynka}}, \bibinfo {author} {\bibfnamefont {A.}~\bibnamefont {Saxe}}, \
  and\ \bibinfo {author} {\bibfnamefont {L.}~\bibnamefont {Saglietti}},\ }\href
  {\doibase 10.1088/1742-5468/ad864b} {\bibfield  {journal} {\bibinfo
  {journal} {Journal of Statistical Mechanics: Theory and Experiment}\ }\textbf
  {\bibinfo {volume} {2024}},\ \bibinfo {pages} {114001} (\bibinfo {year}
  {2024})}\BibitemShut {NoStop}%
\bibitem [{\citenamefont {Varoquaux}\ \emph {et~al.}(2024)\citenamefont
  {Varoquaux}, \citenamefont {Luccioni},\ and\ \citenamefont
  {Whittaker}}]{varoquaux2024hype}%
  \BibitemOpen
  \bibfield  {author} {\bibinfo {author} {\bibfnamefont {G.}~\bibnamefont
  {Varoquaux}}, \bibinfo {author} {\bibfnamefont {A.~S.}\ \bibnamefont
  {Luccioni}}, \ and\ \bibinfo {author} {\bibfnamefont {M.}~\bibnamefont
  {Whittaker}},\ }\href@noop {} {\bibfield  {journal} {\bibinfo  {journal}
  {arXiv:2409.14160}\ } (\bibinfo {year} {2024})}\BibitemShut {NoStop}%
\bibitem [{\citenamefont {Evci}\ \emph {et~al.}(2022)\citenamefont {Evci},
  \citenamefont {van Merriënboer}, \citenamefont {Unterthiner}, \citenamefont
  {Vladymyrov},\ and\ \citenamefont {Pedregosa}}]{evci2022GradMax}%
  \BibitemOpen
  \bibfield  {author} {\bibinfo {author} {\bibfnamefont {U.}~\bibnamefont
  {Evci}}, \bibinfo {author} {\bibfnamefont {B.}~\bibnamefont {van
  Merriënboer}}, \bibinfo {author} {\bibfnamefont {T.}~\bibnamefont
  {Unterthiner}}, \bibinfo {author} {\bibfnamefont {M.}~\bibnamefont
  {Vladymyrov}}, \ and\ \bibinfo {author} {\bibfnamefont {F.}~\bibnamefont
  {Pedregosa}},\ }\href@noop {} {\bibfield  {journal} {\bibinfo  {journal}
  {arXiv:2201.05125}\ } (\bibinfo {year} {2022})}\BibitemShut {NoStop}%
\end{thebibliography}%

\end{document}